\documentclass[twoside,leqno,twocolumn]{article}

\usepackage[letterpaper]{geometry}

\usepackage{ltexpprt}
\usepackage{hyperref}
    
\usepackage{algorithm}
\usepackage{algpseudocode}
\algrenewcommand\algorithmicrequire{\textbf{Input:}}
\algrenewcommand\algorithmicensure{\textbf{Output:}}
\algnewcommand\algorithmicinput{\textbf{Parameter:}}
\algnewcommand\Parameter{\item[\algorithmicinput]}

\usepackage{graphicx}
\usepackage{subcaption}
\usepackage{booktabs}
\usepackage{multirow}
\usepackage{xspace}
\usepackage{color}
\usepackage{enumitem}
\usepackage{threeparttable}

\newcommand{\para}[1]{\underline{\smash{\emph{#1}}}}
\newcommand{\methodlong}[0]{\underline{T}ransition M\underline{a}trix \underline{R}epresentation with \underline{T}ransposed Convolutions\xspace}
\newcommand{\method}[0]{\textsc{Tart}\xspace}

\usepackage{amsmath}
\usepackage{amssymb}
\DeclareMathOperator*{\argmax}{arg\,max}

\newtheorem{definition}{Definition}

\begin{document}

\newcommand\relatedversion{}

\title{\Large Transition Matrix Representation of Trees with Transposed Convolutions}
\author{Jaemin Yoo\thanks{Seoul National University (jaeminyoo@snu.ac.kr).}
\and Lee Sael\thanks{Corresponding author. Ajou University (sael@ajou.ac.kr).}}

\date{}

\maketitle


\fancyfoot[R]{\scriptsize{Copyright \textcopyright\ 2022 by SIAM\\
Unauthorized reproduction of this article is prohibited}}






\begin{abstract} \small\baselineskip=9pt 
How can we effectively find the best structures in tree models?
Tree models have been favored over complex black box models in domains where interpretability is crucial for making irreversible decisions.
However, searching for a tree structure that gives the best balance between the performance and the interpretability remains a challenging task.
In this paper, we propose \method (\methodlong), our novel generalized tree representation for optimal structural search. 
\method represents a tree model with a series of transposed convolutions that boost the speed of inference by avoiding the creation of transition matrices.
As a result, \method allows one to search for the best tree structure with a few design parameters, achieving higher classification accuracy than those of baseline models in feature-based datasets.
\end{abstract}

\section{Introduction} 

Tree models \cite{BreimanFOS84} have been favored over complex black-box models in domains where interpretability is a crucial factor for making reliable decisions, such as in biological and medical fields~\cite{Che11, Jalali12}, where decisions make irreversible effects.
The main advantage of tree models over other classifiers is that their decision processes are understandable without post-processing methods \cite{LIME, GradCAM} that explain approximate reasons for decisions.

Recent works improve the performance of tree models by adopting complex decision functions~\cite{Frosst17, Yoo19, Yoo21} or utilize tree-structured decisions as a component of large black-box models to gain in interpretability~\cite{KontschiederFCB15, RoyT16, 0002GWZWY18}.
However, although these approaches have fundamental similarities in dealing with trees, there is no unified way to generalize and represent them by a single framework.
This makes one resort to manually search for the best tree structure only among a few feasible choices, losing the opportunity to improve in performance.

In this work, we propose \method (\methodlong), a novel framework for generalizing tree models with a unifying view. 
\method characterizes a tree model as a sequence of linear transformations whose transition matrices are determined by input features.
The unified representation of trees allows us a) to effectively characterize and categorize existing models and b) to perform a systematic search over possible tree structures.
\method also utilizes transposed convolutions to avoid the generation of large transition matrices during inference.
This optimization improves the speed of training and inference especially in trees with large depth.

We perform extensive experiments on 121 feature-based datasets and show that \method outperforms existing classifiers with reasonable choices of the tree structure.
We also provide detailed guidelines on the design choices of \method by thorough comparisons between different combinations of parameters.

Our contributions are summarized as follows:
\begin{itemize}[noitemsep]
    \item \textbf{General representation.}
    	We propose \method, a general and efficient tree representation that gives a unifying view of existing tree models.
    \item \textbf{Categorization and characterization.}
   		We analyze existing tree and non-tree classifiers based on the generalizability of \method.
	\item \textbf{Ablation study.}
		We undergo extensive ablation study on 121 tabular datasets to analyze the effects of the design parameters of \method.
	\item \textbf{Efficiency.}
		\method speeds up the inference of tree models up to 36.3 times based on the utilization of transposed convolution operations.
\end{itemize}

The rest of this paper is organized as follows.
We review related works in Section \ref{sec:related-works} and propose \method in Section \ref{sec:framework_GRT}.
We discuss how \method generalizes existing classifiers in Section \ref{sec:structures}.
We present experimental results in Section \ref{sec:exp} and conclude at Section \ref{sec:conclusion}. 
Symbols used in this paper are summarized as Table \ref{table:symbols}.
Our code is available at \underline{\smash{\url{https://github.com/leesael/TART}}}.

\begin{figure*}
	\centering
	\begin{subfigure}{0.43\textwidth}
		\includegraphics[width=\textwidth]{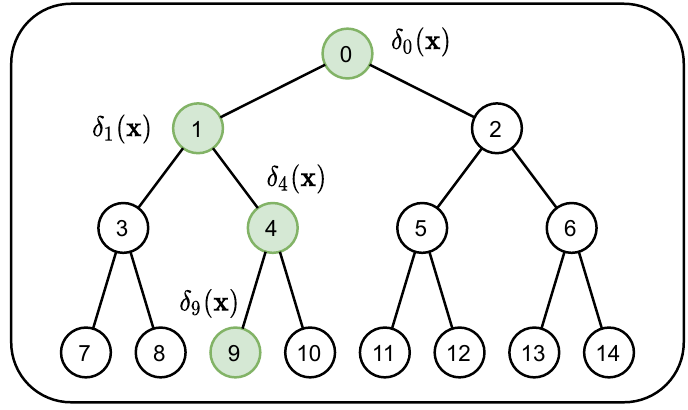}
		\caption{A traditional view of a tree model.}
		\label{fig:motivation-1}
	\end{subfigure} \hfill
	\begin{subfigure}{0.54\textwidth}
		\includegraphics[trim=0.9cm 0 0.6cm 0, clip, width=\textwidth]{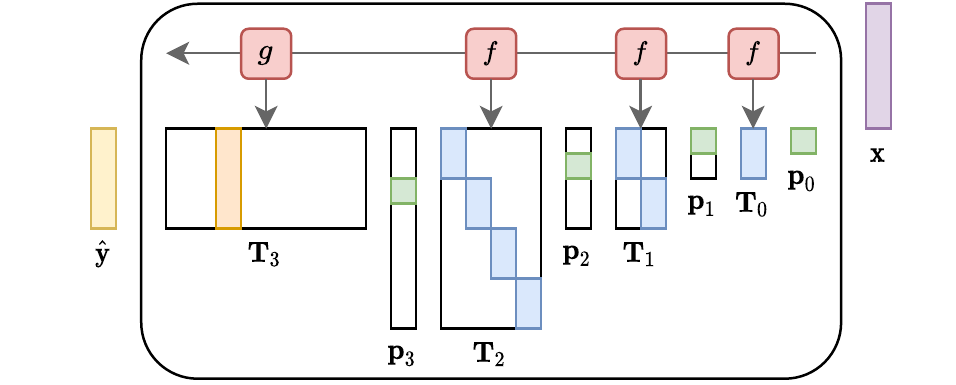}
		\caption{A \method view of a tree model.}
		\label{fig:motivation-2}
	\end{subfigure}
	\caption{%
		The illustration of a binary tree model by (a) the traditional node-and-branch view and (b) the view of our \method utilizing transition matrices.
		The traditional view treats the prediction $\hat{\mathbf{y}}$ as a sequence of decisions $\delta_0$, $\delta_1$, $\delta_4$, and $\delta_9$, while the view taken by our \method treats the model as a series of linear transformations where each matrix $\mathbf{T}_d$ represents the transition at each depth $d \in \{0, 1, 2, 3\}$ from the root to the leaf nodes.
	}
\label{fig:motivation}
\end{figure*}

\section{Related Works} \label{sec:related-works}

Decision trees (DT) propagate input data from the root to the leaf nodes through tree-structured layers without updating their representations~\cite{BreimanFOS84}.
This process is considered inherently interpretable, but typical DTs often show poor performance due to the low generalizability to unseen test data.
We list three types of related works that focus on improving the accuracy of DTs.

\para{Tree models with linear decisions.}
Soft decision trees (SDT) \cite{Irsoy12} are characteristic in that the internal decisions are made by logistic classifiers that utilize all elements of each feature vector.
The logistic classifiers allow differentiable updates of the parameters in SDTs through backpropagation.
SDTs have been studied and used widely due to their simplicity and generalizability \cite{Frosst17, IrsoyA15, linero2018bayesian, Yoo19}.
Deep neural decision trees \cite{Yang18a} extend DTs into multi-branched trees by splitting each example directly into multiple bins using a set of learnable thresholds.
These models have an interpretable nature due to the linearity of decision and leaf functions.

\begin{table}
	\small
    \centering
    \caption{%
        Symbols frequently used in this paper.
    }
    \begin{tabular}{c|l}
        \toprule
        \textbf{Symbol} & \textbf{Description} \\
        \midrule
        $\mathbf{T}_d$ &
        	Transition matrix at layer $d$ \\
        $\mathbf{p}_d$ &
        	Assignment vector at layer $d$ \\
		$f$ & Internal decision function \\
		$g$ & Leaf classifier function \\
		$h$ & Leaf-combining function \\
        \midrule
        $D$ & Tree depth \\
        $W$ & Window size of convolutions \\
        $S$ & Stride of convolutions \\
        $H$ & Number of layers in $f$ \\
        $L$ & Number of layers in $g$ \\
        \bottomrule
    \end{tabular}
    \label{table:symbols}
\end{table}

\para{Tree models on learned representations.}
Recent works have utilized deep neural networks to provide the ability of representation learning to tree models.
They first learn a better representation of each example using a complex black-box model and use the learned representation as input to tree models instead of the raw features.
One popular approach is to use abstract representations generated from convolutional neural networks \cite{ShenZGY17, 0002GWZWY18, abs-2004-00221} or multilayer perceptrons \cite{BuloK14}.
Such approaches make higher accuracy than those of linear tree models, however, they provide interpretability only on top of the abstract representations.
Thus, the direct relationship between the raw features and predictions is unclear due to the nonlinear feature extraction.

\para{Tree models for data categorization.}
Another approach to combine DTs with deep neural networks is to categorize raw examples by hierarchical decisions before feeding them into black-box classifiers \cite{Murthy16}.
Recent works improve the accuracy of deep neural networks by inserting hierarchical decisions as differentiable operations into a deep neural network, instead of building a complete tree model \cite{MurdockLZD16, McGillP17, BrustD19, TannoAACN19}.
These approaches take advantage of DTs with respect to data clustering, rather than focusing on making interpretable decisions, to improve the decision boundaries learned by black-box learners while minimizing the complexity.

In this work, we focus on generalizing and improving complete tree structures that do not change the input features, which often have associated context information that is useful for interpretation.

\section{Proposed Method} \label{sec:framework_GRT}

We propose \method (\methodlong), a unified approach to represent tree models with a series of transition matrices efficiently with transposed convolutions.
Figure \ref{fig:motivation} shows the transition matrix view of a binary tree, on which our \method is based.
Algorithm \ref{algorithm:method} summarizes the decision process of \method for an input feature vector $\mathbf{x}$, which we explain in detail in Section \ref{ssec:training-and-inference}.

\subsection{Transition Matrix Representation} \label{subsec:representation}

We introduce the transition matrix representation of a tree.
We first define a transition matrix in Definition \ref{def:transition-matrix} and describe its properties in Lemmas \ref{lemma:stochastic-decision-1} and \ref{lemma:stochastic-decision-2}.

\begin{definition} \label{def:transition-matrix}
	A rectangular matrix $\mathbf{T}$ is a transition matrix if $\mathbf{T} \geq 0$ and $\sum_i T_{ij} = 1$ for all $j$.
	We represent the set of all possible transition matrices as $\mathcal{P}$.
\end{definition}

\noindent Following from Definition \ref{def:transition-matrix}, every probability vector $\mathbf{p}$ such that $\mathbf{p} \geq 0$ and $\sum_i p_i = 1$ satisfies $\mathbf{p} \in \mathcal{P}$, since it can be thought of as a matrix of size $|\mathbf{p}| \times 1$.

\begin{lemma}
	Given a matrix $\mathbf{T} \in \mathcal{P}$ of size $l \times m$ and a vector $\mathbf{p} \in \mathcal{P}$ of length $m$, $\mathbf{T} \mathbf{p} \in \mathcal{P}$.
\label{lemma:stochastic-decision-1}
\end{lemma}

\begin{proof}
	Let $\mathbf{q} = \mathbf{T} \mathbf{p}$.
	Then, the following holds:
	\begin{equation*}
		\sum_i q_i =
			\sum_i \sum_j T_{ij} p_j =
			\sum_j p_j \sum_i T_{ij} =
			1.
	\end{equation*}
	Thus, the resulting $\mathbf{q}$ is a probability vector.
\end{proof}

\begin{lemma}
	Given two matrices $\mathbf{T} \in \mathcal{P}$ and $\mathbf{U} \in \mathcal{P}$ of sizes $l \times m$ and $m \times n$, respectively, $\mathbf{T} \mathbf{U} \in \mathcal{P}$.
\label{lemma:stochastic-decision-2}
\end{lemma}

\begin{proof}
	Let $\mathbf{V} = \mathbf{T} \mathbf{U}$.
	Then, for every $j$,
	\begin{equation*}
		\sum_i V_{ij} =
			\sum_i \sum_k T_{ik} U_{kj} =
			\sum_k U_{kj} \sum_i T_{ik} =
			1.
	\end{equation*}
	Thus, the resulting $\mathbf{V}$ is a transition matrix.
\end{proof}

\begin{algorithm}[t]
\caption{\method}
\begin{algorithmic}[1]
	\Require Feature vector $\mathbf{x}$
	\Ensure Prediction $\hat{\mathbf{y}}$
	\Parameter Tree depth $D$,
	 internal decision function $f$,
	 leaf classifier $g$, and
	 leaf-combining function $h$
	\For{each $d \in [0, D)$}
		\State $N_d \gets $ Get the number of nodes at layer $d$
		\State $\mathbf{B}_d \gets \mathrm{Stack}(\{f(\mathbf{x}; \theta_{di}) \mid i \in [1, N_d] \})$
	\EndFor
	\State $\mathbf{p}_0 \gets \mathbf{1}$ \Comment{Vector of length 1}
	\State $\mathbf{p}_D \gets \mathbf{B}_D * (\mathbf{B}_{D-1} * \cdots * (\mathbf{B}_1 * \mathbf{p}_0))$ \Comment{Alg. \ref{algorithm:transposed-conv}}
	\State $\hat{\mathbf{y}} \gets h(\mathbf{p}_D, \{g(\mathbf{x}; \theta_i)\}_{i=1, \cdots, N_D})$ \Comment{Eq. \eqref{eq:multi-selection} or  \eqref{eq:single-selection}}
\end{algorithmic}
\label{algorithm:method}
\end{algorithm}

Given an input feature $\mathbf{x}$, the soft down spread of $\mathbf{x}$ from the root to leaves is represented as a set $\{\mathbf{p}_d\}_d$ of assignment vectors, where $\mathbf{p}_d \in \mathcal{P}$ is for each layer $d$. 
Each node in a layer $d$ computes a decision probability for passing $\mathbf{x}$ to its child node $k$ based on a decision function $f(\mathbf{x})_k$ that sums to one over all $k$s.
This process can be understood as the multiplication of a transition matrix $\mathbf{T}_d \in \mathcal{P}$ and the assignment vector $\mathbf{p}_d$, where $\mathbf{T}_d$ is generated from applying $f$ to all nodes in layer $d$ and combining their outputs.
Based on this, we define the \emph{transition matrix representation} as Definition \ref{def:trees}.

\begin{definition}
	The transition matrix representation of a tree classifier $\mathcal{M}$ is given as
	\begin{equation} \label{eq:method-1}
		\mathcal{M}(\mathbf{x})
			= \mathbf{T}_D \cdots \mathbf{T}_1 \mathbf{T}_0 \mathbf{p}_0,
	\end{equation}
	where $\mathbf{p}_0 = 1$ is the arrival probability to the root node, and $D$ is the tree depth.
	$\mathbf{T}_d \in \mathcal{P}$ is the transition matrix at layer $d$, generated by a decision function $f$ as
	\begin{equation}
		T_{dji} = f(\mathbf{x}; \theta_{di})_j,
		\label{eq:transition}
	\end{equation}
	where $T_{dji}$ refers to the $(j, i)$-th element of $\mathbf{T}_d$, and $\theta_{di}$ is the set of parameters for node $i$ at layer $d$.
\label{def:trees}
\end{definition}

\begin{lemma}
	$\mathcal{M}(\mathbf{x}) \in \mathcal{P}$ for any $\mathbf{x}$.
\end{lemma}

\begin{proof}
	$\mathcal{M}$ is a series of liner transformations done with transition matrices.
	Since $\mathbf{p}_0 \in \mathcal{P}$ in Equation \eqref{eq:method-1}, the lemma is proved due to Lemma \ref{lemma:stochastic-decision-1}.
\end{proof}

\begin{algorithm}[t]
\caption{TConv}
\begin{algorithmic}[1]
	\Require Local transition matrix $\mathbf{B}_d$ of size $W \times N_d$ and arrival probability $\mathbf{p}_d$ at layer $d$
	\Ensure Arrival probability $\mathbf{p}_{d+1}$ of layer $d+1$
	\Parameter Stride $S$
	
	\State $\mathbf{p}_{d+1} \gets \mathbf{0}$ \Comment{Initialize the output}
	\State $j \gets 0$ \Comment{Starting index of an output node}
	\For{each $i \in [1, N_d]$}
		\State $p_{d+1, j:j + W} \gets p_{d+1, j:j + W} + p_{d, i} \mathbf{b}_{d, i}$
		\State $j \gets j + S$
	\EndFor
\end{algorithmic}
	\label{algorithm:transposed-conv}
\end{algorithm}

Figure \ref{fig:motivation} visualizes a binary tree by the traditional view and by the transition matrix representation.
Figure \ref{fig:motivation-1} treats the model as a series of independent decisions following the path of $\mathbf{x}$, while Figure \ref{fig:motivation-2} represents the model as a series of linear transformations.
We denote the decision function of the last layer by $g$, since it is defined differently from the internal decision function $f$ in many tree models.
For example, in decision trees, $g$ is a fixed one-hot vector, while $f$ is a decision function that takes $\mathbf{x}$ as an input.
$\mathbf{T}_D \in \mathcal{P}$ is still satisfied with a different $g$ if we assume a classification task.

The figure also indicates that the nonzero elements of each transition matrix determine the tree shape.
For example, the transition matrices of a binary tree (shown in Figure \ref{fig:motivation-2}) have nonzero values at the block-diagonal positions.
Any tree structure can be represented based on the positions of the nonzero elements in transition matrices that derive from diverse decision function $f$.
We present in Section \ref{sec:structures} the structural generalization of \method for representing existing classifiers.

\subsection{Optimization by Transposed Convolutions}
\label{ssec:gen-transposed-conv}

The transition matrices allow \method to represent general tree structures. 
However, generating the transition matrix $\mathbf{T}_d$ for every layer $d$ requires a heavy computation, e.g., size for $T_d$ is $2^{d+1} \times 2^d$ in a binary tree model.
The overall complexity is $O(2^{2D-1})$ in a binary tree of depth $D$, which is infeasible with large $D$.

We propose to utilize transposed convolutions~\cite{DumoulinV16} in the formation of tree structures to avoid the generation of complete transition matrices in \method.
A transposed convolution maps each input node into multiple output nodes by sliding a small kernel.
Thus, the transposed convolution can be applied to spread input data to child nodes in a tree structure. 
In the rest of this paper, we denote a transposed convolution as TConv for brevity.

\begin{figure}
	\centering
	\begin{subfigure}{0.20\textwidth}
		\includegraphics[width=\textwidth, trim=0 29 32 0, clip]{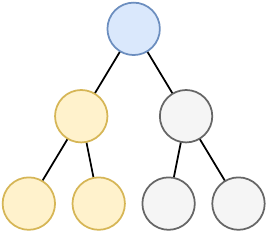}
		\caption{TConv ($W=S=2$)}
		\label{fig:structure-1}
	\end{subfigure} \hspace{2mm}
	\begin{subfigure}{0.254\textwidth}
		\includegraphics[width=\textwidth, trim=0 29 42 0, clip]{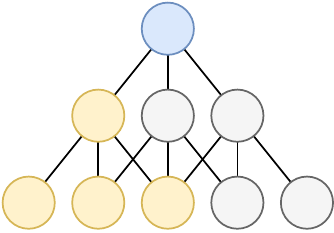}
		\caption{TConv ($W=3, S=1$)}
		\label{fig:structure-2}
	\end{subfigure}

	\caption{%
		Comparison between tree structures determined by the values of $W$ and $S$.
		Intersections between adjacent decisions occur when $S < W$.
	}
	\label{fig:structure}
\end{figure}

Specifically, TConv is utilized in \method as follows.
We are given the arrival probability $\mathbf{p}_d$ of layer $d$ and a decision function $f$.
Then, we create a local transition matrix $\mathbf{B}_d \in \mathbb{R}^{W \times N_d}$ by stacking the outputs of $f$ for all nodes in layer $d$, where $N_d$ is the number of nodes and $W$ is the number of children that each node connects to.
$\mathbf{B}_d$ is then spread out to the assignment vector $\mathbf{p}_{d+1}$ of the next layer by the transposed convolution.
In typical $n$-way trees, $\mathbf{B}_d$ is $n^{d-1}$ times smaller than $\mathbf{T}_d$, allowing us to save extensive time and space in computation.

TConv is then applied to $\mathbf{B}_d$ as described in Algorithm \ref{algorithm:transposed-conv}.
It generates the new arrival probability $\mathbf{p}_{d+1}$ without explicitly generating $\mathbf{T}_d$, given two parameters $W$ and $S$ that determine the shape of the tree.
The kernel slides from the leftmost node in $\mathbf{p}_d$ to the rightmost one, generating $\mathbf{p}_{d+1}$, which is $\in \mathcal{P}$ by Lemma \ref{lemma:stochastic-decision-1}, since $\mathbf{B}_d$ is a transition matrix generated from $f$.

The window size $W$ and the stride $S$ of convolutions are two parameters that determine the shape of a tree.
The window size $W$ determines the branching factor of trees, e.g., $W=2$ in binary trees.
Large $W$ increases the complexity of the decision function $f$ but decreases the tree depth required to make the same number of leaf nodes.
Thus, the value of $W$ makes a tradeoff between the width and depth, and its optimal value depends on the property of $f$ and the characteristic of the dataset.
The stride $S$ determines the number of nodes that are skipped between convolution operations.
Branches have no shared children if $S = W$, since a node slides by the width of the previous decision.
If $S < W$, a node slides less than the width of the previous decision, making a child node take inputs from multiple parents.

\begin{figure}
	\centering
	\includegraphics[width=0.47\textwidth, trim=10 0 10 8, clip]{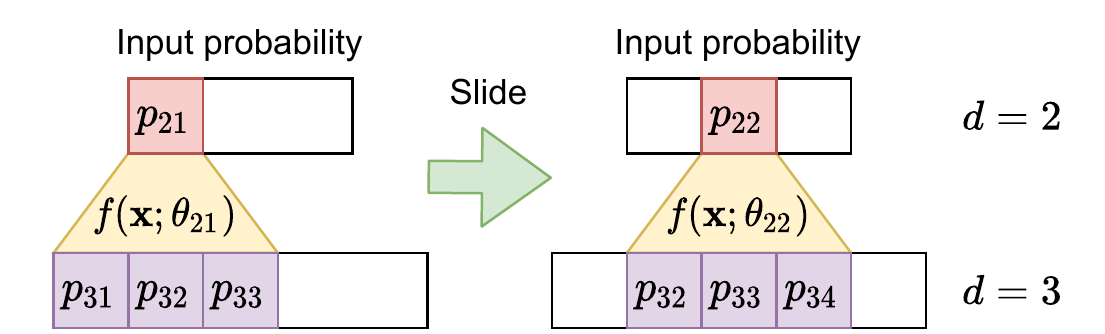}
	\caption{%
		TConv at work between depth 2 and 3 in the tree of Figure \ref{fig:structure-2}.
		The kernel has the width $W=3$ and slides by $S=1$ from the left to the right.
	}
	\label{fig:conv}
\end{figure}

Figure \ref{fig:structure} compares two structures of trees based on the values of $W$ and $S$.
Figure \ref{fig:structure-1} depicts the structure of a typical binary tree, where each node is connected to two child nodes without intersections.
Figure \ref{fig:structure-2} shows a 3-way tree, where each node has three children.
There are two child nodes shared between adjacent decisions since $W - S = 2$.
Figure \ref{fig:conv} is an illustration of TConv in the tree of Figure \ref{fig:structure-2}, when a convolution kernel slides from node 1 to node 2 at layer 2.

\subsection{Training and Inference} \label{ssec:training-and-inference}

We describe how to train \method and how to make its predictions.
We train \method in an end-to-end fashion, updating all parameters by gradient-based optimization.
The objective function is defined as the sum of all loss values from leaf nodes weighted by the arrival probability $\mathbf{p}_D$:
\begin{equation}
	\mathcal{L}(\mathbf{x}, \mathbf{y}) = \sum_{u=1}^{N_D} p_D(u) l(g(\mathbf{x}; \theta_u), \mathbf{y}),
	\label{eq:ens-loss}
\end{equation}
where $g$ is the leaf classifier parameterized with $\theta_u$, $p_D(u)$ is the arrival probability for $u$, and  $l(\hat{\mathbf{y}}, \mathbf{y})$ is the cross entropy function.  The cross entropy $l(\hat{\mathbf{y}}, \mathbf{y})$ is defined as $- \sum_{v \in \mathcal{S}} y(v) \log \hat{y}(v)$, where $\mathcal{S}$ is the set of target classes, $\hat{\mathbf{y}}$ is the prediction, and $\mathbf{y}$ is the true one-hot label vector.

There are two ways to make a decision after \method is trained: a) making a weighted average of predictions from the leaf nodes by $\mathbf{p}_D$, and b) choosing the leaf node that gives the largest arrival probability.
We call these two choices \emph{multi-leaf selection} and \emph{single-leaf selection}, respectively.
The multi-node selection produces higher accuracy in general, resembling ensemble learning, while the single-leaf selection is better for interpretability as a single leaf node participates in each prediction.

\para{Multi-leaf selection.}
The prediction with the multi-leaf selection is defined as follows:
\begin{equation}
	\mathcal{M}(\mathbf{x}_i) = \sum_{u=1}^{N_D} p_D(u) g(\mathbf{x}_i; \theta_u).
\label{eq:multi-selection}
\end{equation}
In this way, a decision process resembles the weighted ensemble of weak classifiers, which are the leaf nodes in our case.
A model can make accurate predictions even though the representation power of each classifier is not sufficient, due to the effect of ensemble learning.

\para{Single-leaf selection.}
The prediction with the single-leaf selection is defined as follows:
\begin{equation}
	\mathcal{M}(\mathbf{x}_i) = g(\mathbf{x}_i; \theta_{u^*}),
\label{eq:single-selection}
\end{equation}
where $u^* = \argmax_u p_D(u)$ is the leaf node that makes the largest arrival probability among all leaves.
In this way, the ability to split examples to proper leaves plays a crucial role for achieving high accuracy.

\para{Overall algorithm.}
The decision process of \method is summarized as Algorithm \ref{algorithm:method}.
In lines 1 to 4, it runs the decision for every internal node and stacks the results of decisions at each layer.
The local transition matrix $\mathbf{B}_d$ of each layer $d$ is used to run inference through the transposed convolution operations in line 6, where $*$ is the TConv function of Algorithm \ref{algorithm:transposed-conv}.
The predictions of leaf nodes are combined in line 7 by the leaf-combining function $h$, based on the arrival probability $\mathbf{p}_D$.

\section{Further Analysis} \label{sec:structures}

We characterize and categorize existing classifiers based on the generalized representation of \method.
We also present three promising combinations of design parameters of \method that have different advantages.

\begin{table}
    \centering
    \caption{%
        Representation of existing binary tree models as \method.
        $f$ and $g$ refer to the internal and leaf decision function, respectively.
        Details are in Section \ref{ssec:gen-binary-trees}.
    }
    \begin{tabular}{l|c|c}
        \toprule
        Model & $f(\mathbf{x}; \theta_i)$ & $g(\mathbf{x}; \theta_j)$ \\
        \midrule
        DT \cite{BreimanFOS84} &
            $\mathbb{I}(s_i (\mathbf{1}_i^\top \mathbf{x} - b_i) > 0)$ &
            $\mathrm{Onehot}(\theta_j)$ \\
        SDT \cite{Irsoy12} &
            $\sigma(\mathbf{w}_i^\top \mathbf{x} + b_i)$ &
            $\mathrm{Categorical}(\theta_j)$ \\
        \midrule
        NDF \cite{BuloK14} &
            $\mathrm{MLP}_i(\mathrm{rand}(\mathbf{x}))$ &
            $\mathrm{Categorical}(\theta_j)$ \\
        DNDF \cite{KontschiederFCB15} &
            $\mathrm{CNN}(\mathbf{x}; i)$ &
            $\mathrm{Categorical}(\theta_j)$ \\
        NRF \cite{RoyT16} &
            $\mathrm{CNN}(\mathbf{x}; i, \mathrm{depth}(i))$ &
            $\mathrm{Gaussian}(\theta_j)$ \\
        \bottomrule
    \end{tabular}
    \label{table:tree-models}
\end{table}

\subsection{Generalizability} \label{ssec:gen-binary-trees}

We study the generalizability of \method in binary trees and general classifiers.

\para{Representation of binary trees.}
Existing tree models have different characteristics but share a similar tree structure.
Such models differ in the choice of decision functions $f$ and $g$ working at the internal layers and the leaf layer, respectively.
We show in Table \ref{table:tree-models} how \method represents different tree models with the choice of $f$ and $g$.
We set the structural parameters $W$ and $S$ to $2$, since all these models have the binary tree structure.

Decision trees (DT) select a single element of each input feature $\mathbf{x}$ by a one-hot vector $\mathbf{1}_i$ and compare it with a learned threshold $b_i$ at each internal node $i$.
Soft decision trees (SDT) improve DTs by performing a soft decision at each branch, which uses all elements of $\mathbf{x}$ as a linear separator using the logistic sigmoid function $\sigma$.
The weight vector $\mathbf{w}_i$ is learned for each node $i$.
Their decision processes are naturally interpretable, since the decision functions are linear with respect to $\mathbf{x}$.

\begin{table}
	\centering
	\caption{%
		Classifier models represented by \method with three design parameters: tree depth D, the number $H$ of layers in $f$, and the number $L$ of layers in $g$.
	}
	\small
	\begin{tabular}{l|cccc}
		\toprule
		Models & $D$ & $H$ & $L$ \\
		\midrule
		Logistic regression & $D = 0$ & - & $L = 1$ \\
		Multilayer perceptrons \cite{Olson18} & $D = 0$ & - & $L > 1$ \\
		\midrule
		Simple ensembles of experts & $D > 0$ & $H = 0$ & Any $L$ \\
		\midrule
		Trees of type 1 \cite{BreimanFOS84, Irsoy12} & $D >0$ & $H = 1$ & $L = 0$ \\
		Trees of type 2 \cite{KontschiederFCB15, RoyT16} & $D > 0$ & $H > 1$ & $L = 1$ \\
		Trees of type 3 \cite{MurdockLZD16, Murthy16} & $D > 0$ & $H = 1$ & $L > 1$ \\
		\bottomrule
	\end{tabular}
\label{table:classifiers}
\end{table}

The remaining models use nonlinear decision functions.
Neural decision forests (NDF) utilize a randomized multilayer perceptron (MLP) as a decision function.
Deep neural decision forests (DNDF) use a single convolutional neural network (CNN) for all decisions, changing only the last fully-connected layer.
Neural regression forests (NRF) and their variants use hierarchical CNNs having different numbers of convolutions \cite{0002GWZWY18, ShenZGY17}.
All of these models use deep neural networks as their decision functions to improve representation power.

\begin{table}
	\setlength\tabcolsep{5.3pt}
	\small
    \centering
    \caption{%
        Promising tree structures of \method that have different properties.
        Details are in Section \ref{ssec:summary-choices}.
    }
    \begin{tabular}{l|cc|ccc|l}
        \toprule
        \textbf{Model} & $W$ & $S$ & $D$ & $H$ & $L$ & Property \\
        \midrule
        \method-A & 2 & 2 & 6 & 1 & 1 & Strong in small data \\
        \method-B & 2 & 2 & 2 & 1 & 4 & Strong in large data \\
        \method-C & 3 & 2 & 3 & 1 & 2 & Best balance \\
        \bottomrule 
    \end{tabular}
\label{table:proposed-models}
\end{table}

\para{Categorization of general classifiers.}
We utilize the framework of \method to categorize and characterize existing classifiers.
We assume that deep neural networks having a nonlinear activation function are used for both $f$ and $g$.
Then, we introduce three design parameters of \method as the main variables: tree depth $D$, the number $H$ of layers in $f$, and the number $L$ of layers in $g$.
The result of categorization is given as Table \ref{table:classifiers}.

A classifier is a single expert having no tree structure if $D=0$.
In this case, logistic regression (LR) and MLPs are distinguished by the value of $L$.
If $H=0$, no internal decisions are made even with $D>1$, meaning that all examples are equally split into all leaf nodes.
In this case, a classifier makes a prediction by computing the simple average of predictions as an ensemble model.
A classifier splits given examples by learnable decisions only if $D>0$ and $H>0$, becoming a tree model whose structure represents a decision path.

The characteristic of a tree classifier is determined by the values of $H$ and $L$.
Models having $H = 1$ and $L=0$ split given examples by linear decisions into leaf classifiers that return fixed predictions.
Thus, they are the simplest tree models that focus on interpretability.
Models with $H>1$ focus on the ability to split examples by utilizing a nonlinear decision function at the internal nodes, while those with $H=1$ and $L>1$ use a simple decision rule but focus on the leaf classifiers.

\subsection{Promising Tree Structures}
\label{ssec:summary-choices}

Based on the categorization of existing models, we propose three promising structures of \method consisting of different values of parameters.
Table \ref{table:proposed-models} summarizes the structures, which we call \method-A, \method-B, and \method-C, respectively.
We assume decision functions $f$ and $g$ as multilayer perceptrons with $H$ and $L$ layers, respectively, as in Table~\ref{table:classifiers}.
We set $H=1$ in this case, because we have found that $H>1$ makes a tree model easily overfit to training data without a clear advantage in our datasets.

\begin{table}
	\setlength\tabcolsep{5.3pt}
	\centering
	\caption{%
		The information of 121 datasets divided into three groups by the number of examples, which include 9, 37, and 75 datasets, respectively.\textsuperscript{1}
	}
	\small
	\begin{threeparttable}
	\begin{tabular}{l|rr|c|c}
		\toprule
		\multirow{2}{*}{\textbf{Group}}
			& \multicolumn{2}{c|}{\textbf{Examples}}
			& \multicolumn{1}{c|}{\textbf{Features}}
			& \multicolumn{1}{c}{\textbf{Labels}} \\
		& \textbf{Min}
			& \textbf{Max}
			& \textbf{Avg $\pm$ Std}
			& \textbf{Avg $\pm$ Std} \\
		\midrule
		Large & 10,992 & 130,064 & 19.0 $\pm$ 15.8 & 8.2 $\pm$ 8.5 \\
		Mid   & 1,000  & 8,124   & 40.2 $\pm$ 48.4 & 12.2 $\pm$ 26.6 \\
		Small & 10     & 990     & 24.4 $\pm$ 37.9 & 4.1 $\pm$ 3.7 \\
		\midrule
		All & 10     & 130,064  & 28.8 $\pm$ 40.8 & \ 6.9 $\pm$ 15.5 \\
		\bottomrule
	\end{tabular}
	\begin{tablenotes}
		\item[1] \footnotesize{\url{http://persoal.citius.usc.es/manuel.fernandez.delgado/papers/jmlr}}
	\end{tablenotes}
	\end{threeparttable}
	\label{table:datasets}
\end{table}

\para{Linear leaves (\method-A).}
A tree model with linear decision functions gives clear interpretability.
\method-A is characterized by an abundant number of leaf nodes each of which makes a linear decision boundary for the examples that have arrived through internal decisions.
\method-A performs the best in small datasets, where the nonlinearity is not essential for acquiring high accuracy.
On the other hand, the linearity of \method-A allows one to avoid overfitting in such small datasets, resulting in improving accuracy in unseen test data.

\para{Nonlinear leaves (\method-B).}
The linearity requires us to use a sufficient number of leaf nodes to make high accuracy. 
On the other hand, we can bound the number of leaves if we increase the capacity of each leaf node.
This turns our model into a small ensemble of nonlinear classifiers, where any leaf selection scheme can be used with a different advantage: the single-leaf selection has better interpretability of decisions, while the multi-leaf selection improves performance.
Still, we focus on only the single-leaf selection, as our primary goal of utilizing tree models is to make clear interpretability.

\para{Three-way decisions (\method-C).}
\method-C focuses on the balance between \method-A and \method-B.
Three-way branches with $W=3$ make each internal decision richer than in binary trees.
Still, it makes the width of a tree increases much faster with the tree depth than in binary trees.
Thus, we make intersections between decisions by setting $S=2$ to bound the tree width while utilizing the rich decisions.
The choices of other parameters such as $D$ and $L$ are in between those of \method-A and \method-B.
The chosen structure is similar to Figure \ref{fig:structure-2}, except that \method-C slides the kernel by two instead of one.

\section{Experiments}
\label{sec:exp}

We compare our \method with existing tree and non-tree classifiers by experiments on feature-based data, where tree models have been adopted actively.

\begin{table}
	\small
    \centering
    \caption{%
        Classification accuracy of \method and baseline models.
        MLP-$l$ represents an MLP having $l$ layers.
        Our three \method models show the best accuracy in different groups of datasets, based on their characteristics.
    }
    \begin{tabular}{l|ccc}
        \toprule
        \textbf{Model} &
            \textbf{Large} &
            \textbf{Medium} &
            \textbf{Small} \\
        \midrule
        DT &
            88.3$\pm$0.1 &
            76.3$\pm$0.2 &
            71.9$\pm$0.7 \\
        LR &
            79.1$\pm$0.1 &
            80.8$\pm$0.2 &
            75.8$\pm$0.3 \\
        SVM-lin &
            77.7$\pm$0.1 &
            79.0$\pm$0.2 &
            74.9$\pm$0.5 \\
        SVM-rbf &
            87.6$\pm$0.0 &
            81.1$\pm$0.1 &
            \textbf{77.0$\pm$0.2} \\
        \midrule
        MLP-1 &
            78.7$\pm$0.1 &
            78.9$\pm$0.3 &
            73.4$\pm$0.4 \\
        MLP-2 &
            87.8$\pm$0.1 &
            \underline{83.0$\pm$0.4} &
            76.5$\pm$0.4 \\
        MLP-4 &
            \underline{91.8$\pm$0.1} &
            \underline{83.0$\pm$0.2} &
            76.8$\pm$0.2 \\
        MLP-8 &
            91.5$\pm$0.1 &
            82.5$\pm$0.3 &
            76.0$\pm$0.5 \\
        MLP-16 &
            85.3$\pm$0.9 &
            78.3$\pm$0.2 &
            75.1$\pm$0.6 \\
        \midrule
        \method-A &
            88.2$\pm$0.2 &
            82.6$\pm$0.2 &
            \textbf{77.0$\pm$0.6} \\
        \method-B &
            \textbf{92.1$\pm$0.1} &
            82.7$\pm$0.4 &
            76.0$\pm$0.3 \\
        \method-C &
            89.6$\pm$0.4 &
            \textbf{83.1$\pm$0.2} &
            76.3$\pm$0.1 \\
        \bottomrule
    \end{tabular}
    \label{table:baselines}
\end{table}

\para{Datasets.}
We use 121 feature-based datasets taken from UCI Machine Learning Repository \cite{UCI}, which were used as a benchmark in \cite{Delgado14, Olson18}.
Table \ref{table:datasets} summarizes the information of our datasets, which are categorized into three groups by the number of examples.
We follow the experimental setup of \cite{Olson18} including the data split and feature preprocessing.
In all experiments, we run each model four times with different random seeds and report the average and standard deviation.

\para{Baselines.}
We include the following baseline classifiers in our experiments, which have been used widely for feature-based datasets: logistic regression (LR), decision trees (DT), and support vector machines (SVM) with the linear and RBF kernels.
We also include multilayer perceptrons (MLP) as a strong competitor, whose structure is taken from a previous work that studied our UCI datasets \cite{Olson18}: 100 units at each hidden layer, the ELU activation function \cite{Clevert15}, He-initialization \cite{He15}, and dropout of probability 0.15 \cite{Srivastava14}.
The training of MLPs follows the same process as our \method.

\para{Hyperparameters.}
We adopt an MLP with the same ELU activation and dropout of probability 0.15 as the decision functions $f$ and $g$ of \method.
We train \method and MLPs based on the Adam optimizer \cite{Kingma14} with the initial learning rate 0.005.
The batch size is set to 1024, which is large enough to load most datasets by a single batch.
We ran all of our experiments on a workstation having GTX 1080 Ti, based on PyTorch.
We use classification accuracy as a metric to evaluate all classifiers.

\subsection{Classification Accuracy}
\label{ssec:exp-accuracy}

We compare the accuracy of \method and baseline models in Table \ref{table:baselines}.
Our \method models show the highest accuracy in general, with their strengths in different groups of datasets.

\begin{table}
	\small
    \centering
    \caption{%
        Accuracy of \method when the linear leaf nodes are adopted.
        Models with multi-leaf selection perform better than single-leaf models in most cases, and both models show higher accuracy with larger $D$.
    }
    \begin{tabular}{crr|ccc}
        \toprule
        \textbf{Leaves} &
            \textbf{$D$} &
            \textbf{$L$} &
            \textbf{Large} &
            \textbf{Medium} &
            \textbf{Small} \\
        \midrule
        Multi & 2 & 1 &
            84.6$\pm$0.2 &
            81.4$\pm$0.3 &
            75.3$\pm$0.5 \\
        Multi & 4 & 1 &
            86.7$\pm$0.1 &
            82.1$\pm$0.3 &
            76.2$\pm$0.2 \\
        Multi & 6 & 1 &
            88.2$\pm$0.2 &
            \underline{82.6$\pm$0.2} &
            \textbf{77.0$\pm$0.6} \\
        Multi & 8 & 1 &
            \textbf{89.1$\pm$0.1} &
            \textbf{82.9$\pm$0.4} &
            \underline{76.5$\pm$0.6} \\
        \midrule
        Single & 2 & 1 &
            84.4$\pm$0.2 &
            81.1$\pm$0.3 &
            74.7$\pm$0.5 \\
        Single & 4 & 1 &
            86.4$\pm$0.1 &
            81.6$\pm$0.4 &
            75.1$\pm$0.4 \\
        Single & 6 & 1 &
            87.8$\pm$0.2 &
            82.0$\pm$0.3 &
            75.7$\pm$0.6 \\
        Single & 8 & 1 &
            \underline{88.6$\pm$0.1} &
            82.2$\pm$0.3 &
            74.6$\pm$0.6 \\
        \bottomrule
    \end{tabular}
    \label{table:linear-leaves}
\end{table}

DTs show the lowest accuracy in the medium and small datasets, since they easily overfit to training data.
MLPs and SVM with the RBF kernel perform the best among the baselines due to the nonlinearity of decisions.
MLP-1 works in a similar way to LR, but its accuracy is lower than those of LR and SVM-lin.
This is because the stochastic training of MLPs does not guarantee the global optimum of parameters.
The accuracy of MLPs depends heavily on the number of layers, indicating the sensitivity to the choice of hyperparameters.

Our three \method models show the best accuracy in different groups of datasets.
\method-A works the best in small datasets since it consists of linear leaf nodes each of which has a limited capacity, minimizing the risk of overfitting.
\method-B achieves the best accuracy in large datasets by combining multiple nonlinear leaves based on tree decisions, each of which has the same structure as MLP-4.
\method-C is a balance between \method-A and \method-B, resulting in the best accuracy for medium-sized datasets among all \method models and baselines.

\subsection{Structural Search}
\label{ssec:exp-ablation-study}

The flexibility of our \method allows us to easily search for a suitable structure by the choice of its design parameters.
We categorize possible options of parameters into three groups that correspond to \method-A, \method-B, and \method-C, respectively.

\para{Linear leaves (\method-A).}
Table \ref{table:linear-leaves} performs an ablation study for \method-A by changing the depth $D$ and the leaf selection function $h$.
All these models use linear leaf nodes to maximize the interpretability, which is the main strength of \method-A.
Multi-leaf models perform better than single-leaf models in general, because they make up for the limited capacity of leaf nodes by combining multiple nodes for each prediction.
Still, single-leaf models work better than the linear baselines such as LR or SVM-lin, since they choose a suitable classifier for each example following the tree structure.
It is also notable that both multi- and single-leaf models perform better with larger $D$, without showing a significant drop of its accuracy unlike MLPs of Table \ref{table:baselines}.

\begin{table}
	\small
    \centering
    \caption{%
        Accuracy of \method when nonlinear leaf nodes are adopted.
        They are specialized for large datasets and achieve higher accuracy than those of MLPs (Table \ref{table:baselines}) or \method models with linear leaves (Table \ref{table:linear-leaves}).
    }
    \begin{tabular}{ccc|ccc}
        \toprule
        \textbf{Leaves} &
            \textbf{$D$} &
            \textbf{$L$} &
            \textbf{Large} &
            \textbf{Medium} &
            \textbf{Small} \\
        \midrule
        Single & 2 & 2 & 
            87.4$\pm$0.1 &
            \textbf{82.7$\pm$0.3} &
            \underline{76.0$\pm$0.2} \\
        Single & 4 & 2 & 
            89.0$\pm$0.1 &
            82.6$\pm$0.5 &
            \textbf{76.1$\pm$0.4} \\
        Single & 6 & 2 & 
            90.0$\pm$0.1 &
            82.6$\pm$0.4 &
            \underline{76.0$\pm$0.4} \\
        Single & 8 & 2 & 
            90.7$\pm$0.0 &
            82.6$\pm$0.4 &
            75.4$\pm$0.3 \\
        \midrule
        Single & 2 & 4 & 
            \underline{92.1$\pm$0.1} &
            \textbf{82.7$\pm$0.6} &
            \underline{76.0$\pm$0.3} \\
        Single & 4 & 4 & 
            \textbf{92.3$\pm$0.1} &
            82.2$\pm$0.4 &
            75.7$\pm$0.3 \\
        Single & 6 & 4 & 
            \underline{92.1$\pm$0.1} &
            81.9$\pm$0.1 &
            75.6$\pm$0.4 \\
        Single & 8 & 4 & 
            91.9$\pm$0.1 &
            82.0$\pm$0.2 &
            75.4$\pm$0.3 \\
        \bottomrule
    \end{tabular}
    \label{table:nonlinear-leaves}
\end{table}

\para{Nonlinear leaves (\method-B).}
Adopting nonlinear leaf nodes requires us to choose the single-leaf selection scheme for interpretability.
Table \ref{table:nonlinear-leaves} compares the performance of \method when $L > 1$, changing the tree depth $D$ from 2 to 8, as an ablation study for \method-B.
Trees with $D\geq4$ and $L=4$ achieve the best accuracy in the large datasets compared to MLPs (in Table \ref{table:baselines}) and trees with linear leaf nodes (in Table \ref{table:linear-leaves}).
Models with $L=2$ work well in the medium and small datasets but show limited performance in the large datasets.

The result implies that the representation power of each leaf classifier is an important factor for achieving high accuracy in large datasets.
At the same time, the split of data examples through tree-structured decisions is effective for improving the performance of classification avoiding overfitting.
This is shown well in Table \ref{table:baselines}, where a significant drop of accuracy is observed when a large number of layers are adopted for MLPs.

\para{Multi-way decisions (\method-C).}
Table \ref{table:branch-intersections} performs an ablation study for \method-C, comparing models with multi-way decisions.
We set the stride $S$ of transposed convolutions to $2$ while changing the tree depth $D$ and the window size $W$.
If $W=3$, the number of leaves at each model of depth $D$ is $2^{D+1} - 1$.
Thus, the first four models in Table \ref{table:branch-intersections} have the same number of leaf nodes as the last four models in the table, respectively.

We observe the effect of branching intersections by comparing the models in Tables \ref{table:nonlinear-leaves} and \ref{table:branch-intersections}.
The first four models in Table \ref{table:branch-intersections} work better than the first four models in Table \ref{table:nonlinear-leaves}, even though they have fewer leaves, taking advantage of intersecting branches.
On the other hand, it is observed from the last four models of Table \ref{table:branch-intersections} that the ability to split data to the leaf nodes is limited when $D=1$, even with large $W$, due to the limited capacity that a single decision function can have.

\begin{table}
	\small
    \centering
    \caption{%
        Accuracy of \method with multi-way decisions and the single-leaf selection.
        These models show similar accuracy with the choice of parameters.
    }
    \begin{tabular}{crc|ccc}
        \toprule
        \textbf{$D$} &
            \textbf{$W$} &
            \textbf{$L$} &
            \textbf{Large} &
            \textbf{Medium} &
            \textbf{Small} \\
        \midrule
        1 & 3 & 2 & 
            88.5$\pm$0.1 &
            83.1$\pm$0.3 &
            \textbf{76.5$\pm$0.7} \\
        3 & 3 & 2 & 
            89.6$\pm$0.1 &
            83.1$\pm$0.4 &
            \underline{76.3$\pm$0.4} \\
        5 & 3 & 2 & 
            \underline{90.7$\pm$0.1} &
            82.9$\pm$0.3 &
            76.2$\pm$0.1 \\
        7 & 3 & 2 & 
            \textbf{91.2$\pm$0.1} &
            82.8$\pm$0.4 &
            75.8$\pm$0.3 \\
        \midrule
        1 & 3 & 2 & 
            88.9$\pm$0.2 &
            \textbf{83.3$\pm$0.1} &
            75.7$\pm$0.9 \\
        1 & 7 & 2 & 
            89.5$\pm$0.1 &
            \textbf{83.3$\pm$0.2} &
            75.6$\pm$0.5 \\
        1 & 15 & 2 & 
            90.1$\pm$0.1 &
            83.1$\pm$0.3 &
            75.4$\pm$0.9 \\
        1 & 31 & 2 & 
            90.4$\pm$0.1 &
            83.2$\pm$0.3 &
            75.5$\pm$0.4 \\
        \bottomrule
    \end{tabular}
    \label{table:branch-intersections}
\end{table}

\subsection{Efficiency}
\label{ssec:exp-efficiency}

A notable advantage of \method is the speedup from existing implementations of tree models due to the efficient computation of transposed convolutions.
We compare \method with three public implementations of soft decision trees (SDT) \cite{Frosst17}, which is a special case of \method with $H=1$, $L=0$, $W=S=2$, and the single-leaf selection.
We call the baselines SDT-K, SDT-X, and SDT-E, respectively, following the first letters of their repository names.\footnote{\url{https://github.com/kimhc6028/soft-decision-tree}}\footnote{\url{https://github.com/xuyxu/Soft-Decision-Tree}}\footnote{\url{https://github.com/endymion64/SoftDecisionTree}}
\method and the all baselines are implemented based on the PyTorch framework. 

We make all methods have the same structure and decision function, changing the tree depth $D$ from 8 to 12.
We use the MNIST dataset \cite{MNIST} in this experiment to be on par with other baselines methods.
We consider each $28 \times 28$ image as a $768$-dimensional vector with no structural information \cite{Frosst17}.
The training set has 60,000 examples, while the test set has 10,000 examples.
We use a single GPU of GTX 1080 Ti and set the batch size to 1024 as in the other experiments.

\begin{figure}
	\centering
	\includegraphics[width=0.45\textwidth]{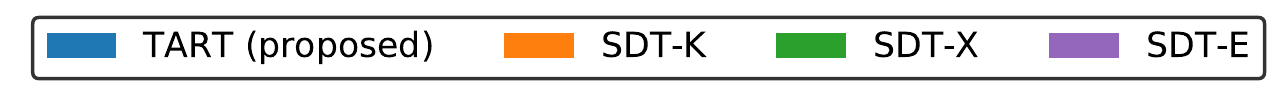}
	
	\begin{subfigure}{0.156\textwidth}
		\includegraphics[width=\textwidth]{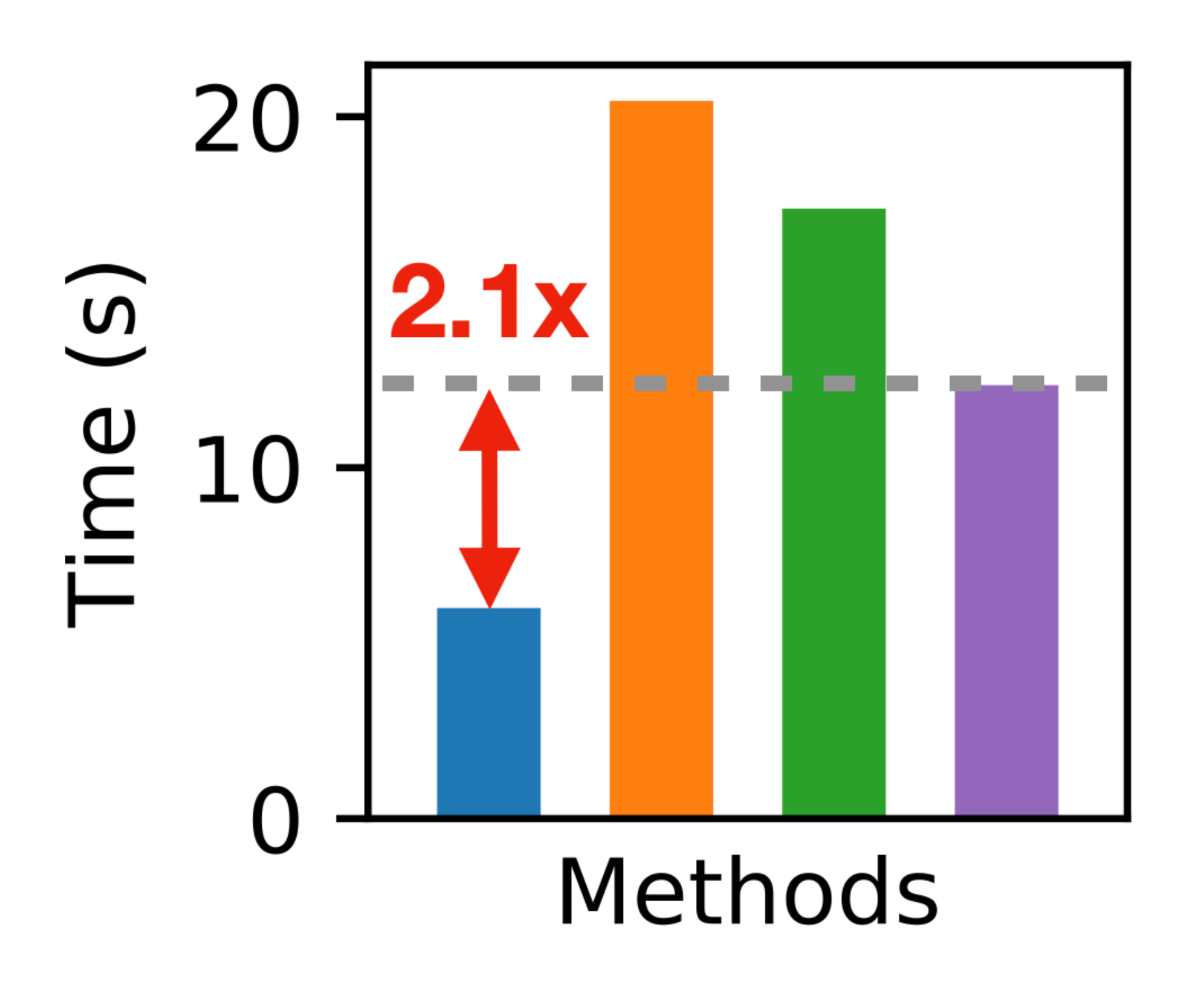}
		\caption{$D=8$.}
		\label{fig:training-speed-1}
	\end{subfigure} \hfill
	\begin{subfigure}{0.156\textwidth}
		\includegraphics[width=\textwidth]{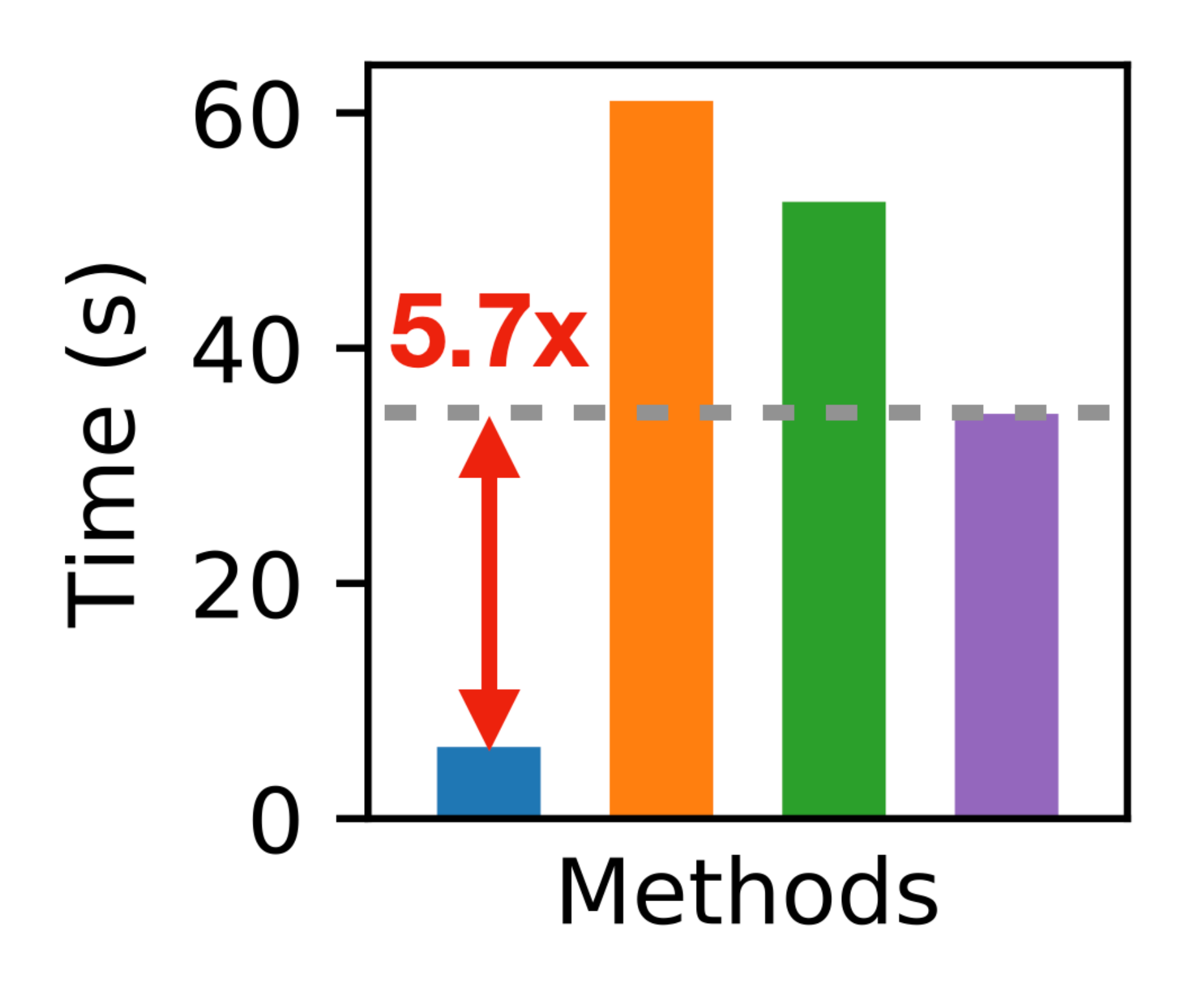}
		\caption{$D=10$.}
		\label{fig:training-speed-2}
	\end{subfigure} \hfill
	\begin{subfigure}{0.1635\textwidth}
		\includegraphics[width=\textwidth]{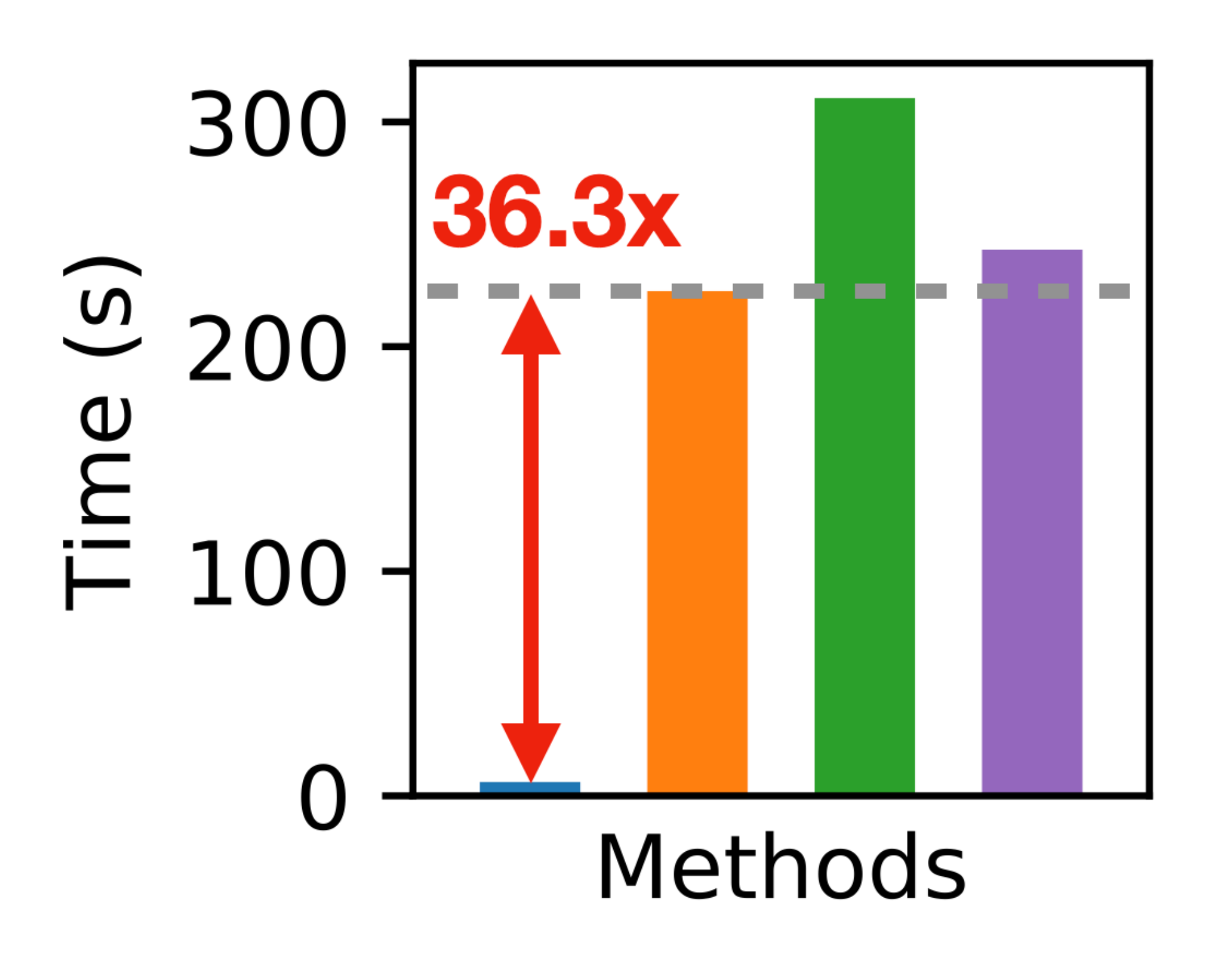}
		\caption{$D=12$.}
		\label{fig:training-speed-2}
	\end{subfigure}

	\caption{%
		Training time of soft decision trees (SDT) by different implementations.
		\method achieves the shortest training time due to its efficiency.
	}
	\label{fig:training-speed}
\end{figure}

\begin{figure}
	\centering
	\includegraphics[width=0.45\textwidth]{figures/bars/legend}
	
	\begin{subfigure}{0.1625\textwidth}
		\includegraphics[width=\textwidth]{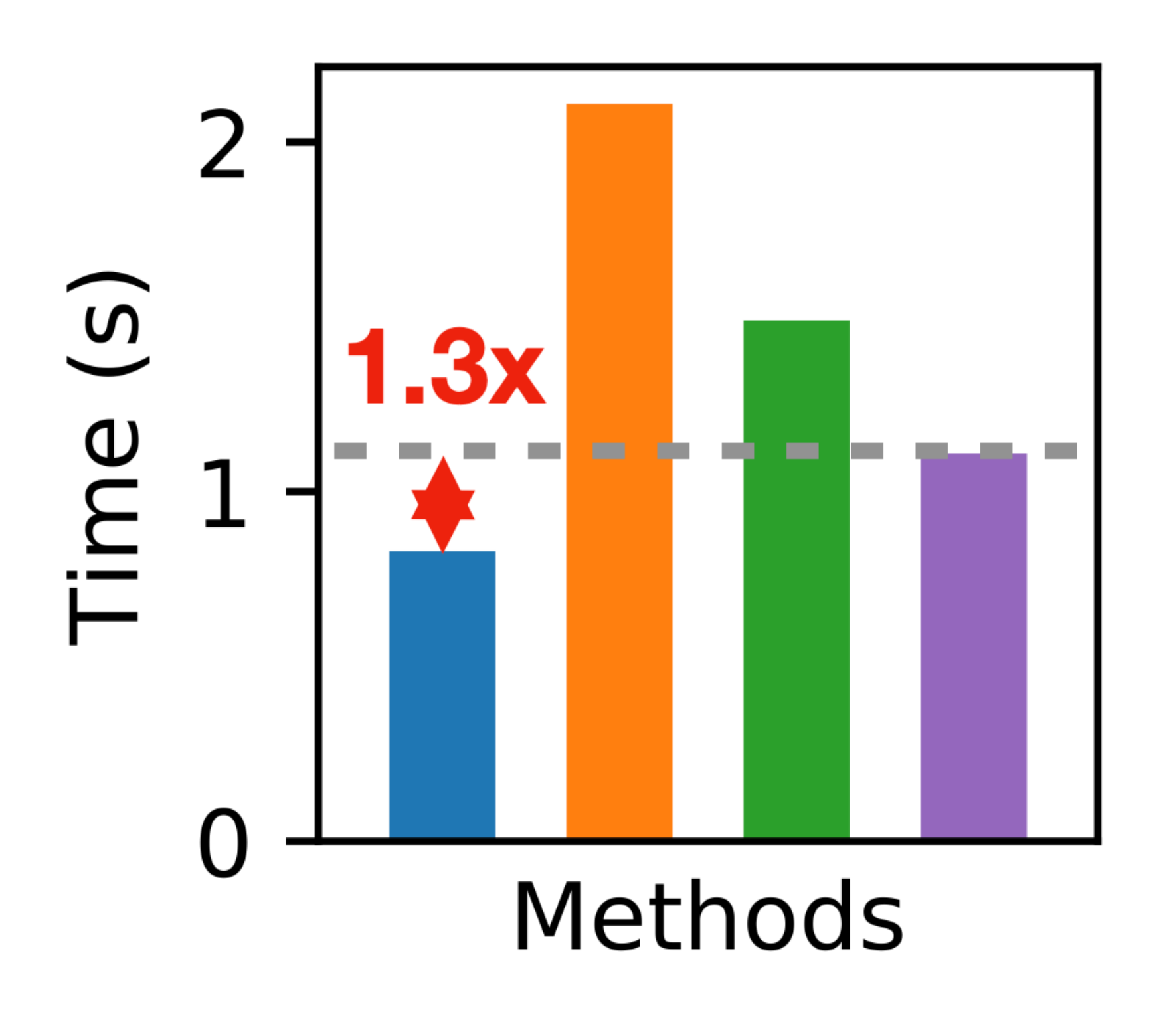}
		\caption{$D=8$.}
		\label{fig:test-speed-1}
	\end{subfigure} \hfill
	\begin{subfigure}{0.151\textwidth}
		\includegraphics[width=\textwidth]{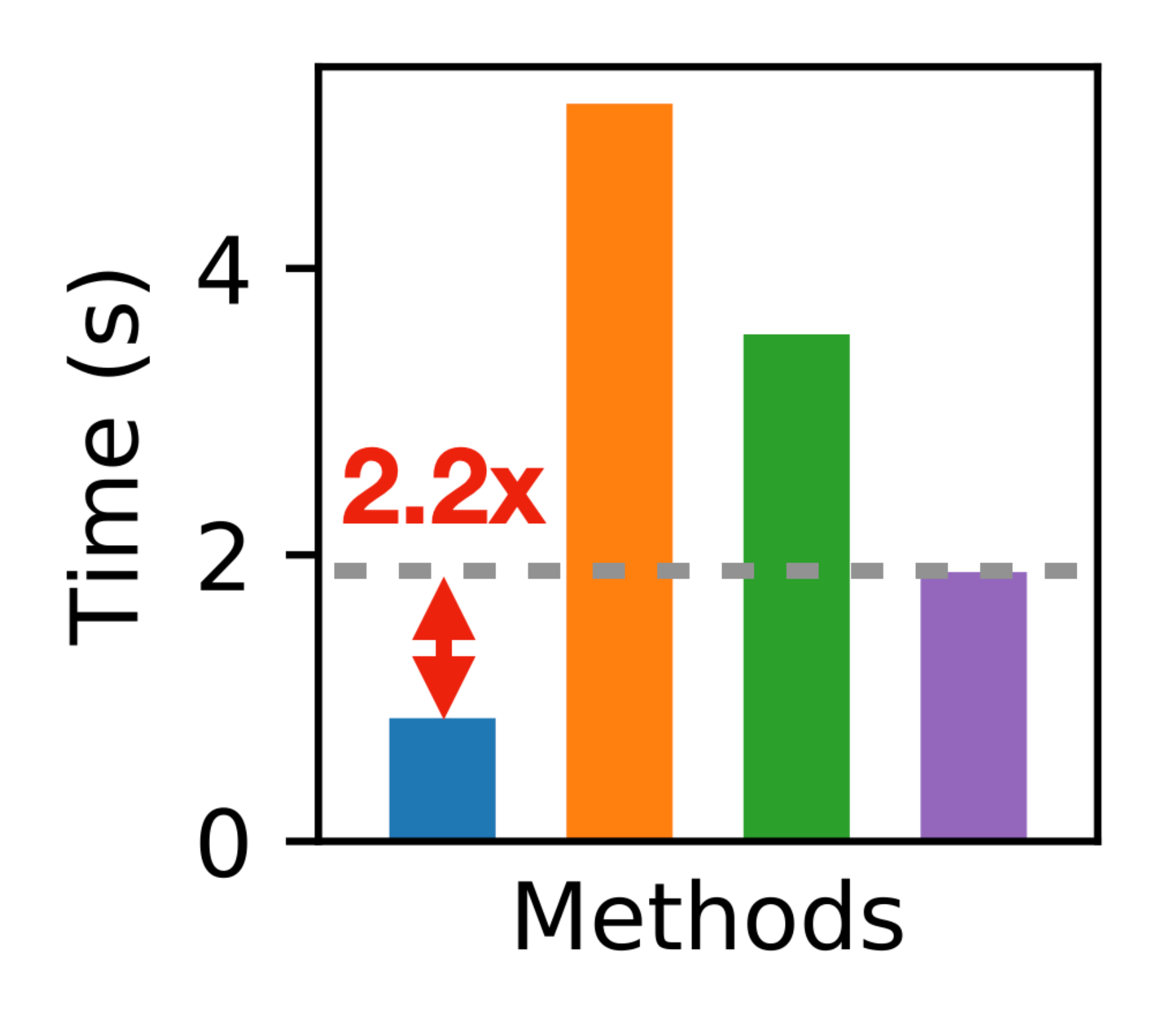}
		\caption{$D=10$.}
		\label{fig:test-speed-2}
	\end{subfigure} \hfill
	\begin{subfigure}{0.159\textwidth}
		\includegraphics[width=\textwidth]{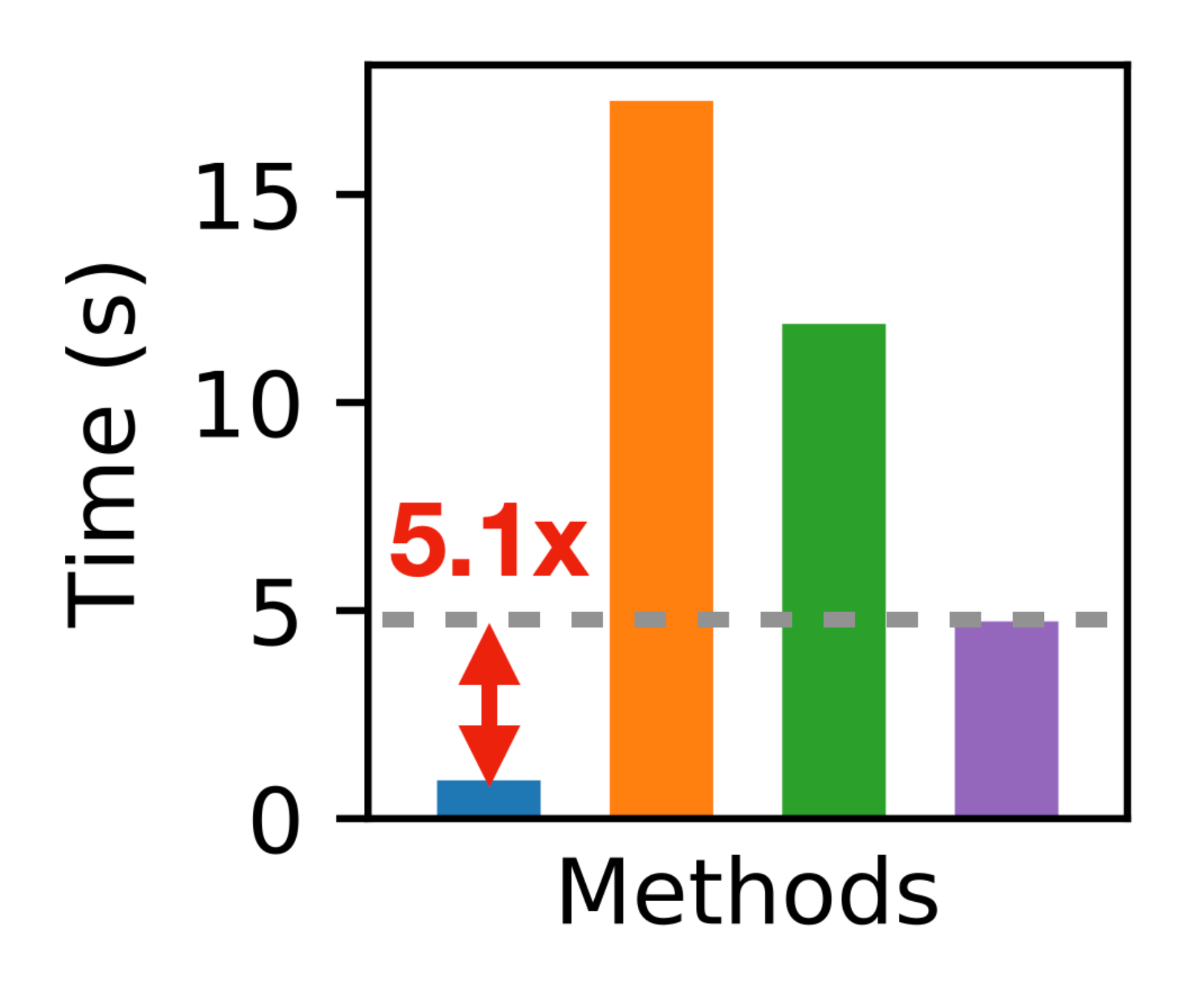}
		\caption{$D=12$.}
		\label{fig:test-speed-2}
	\end{subfigure}

	\caption{%
		Inference time of soft decision trees (SDT) by different implementations.
		\method achieves the shortest inference time due to its efficiency.
	}
	\label{fig:test-speed}
\end{figure}

Figure \ref{fig:training-speed} compares the training time of methods for a single epoch, while Figure~\ref{fig:test-speed} shows the inference time in the test data.
In both experiments, our \method consistently improves the speed of existing implementations.
\method achieves the speedup of up to 36.3$\times$ and 5.1$\times$ in the training and inference, respectively, compared to the best competitors.
This is because the baselines treat a tree model as a set of independent decisions, while \method treats it as a sequence of linear transformations with the efficiency of transposed convolutions.

\section{Conclusion} \label{sec:conclusion}

We propose \method (\methodlong), our novel approach to represent tree models as a series of stochastic decisions efficiently with transposed convolutions.
\method generalizes the structures of different tree models only with a few design parameters.
The generalized representation allows us to systematically search for the best structure for each dataset.
We also present three promising combinations of structural parameters that can be applied to small, medium, and large datasets, respectively.
Our extensive experiments on 121 datasets show that \method achieves the highest accuracy compared to existing classifiers.
At the same time, the optimization with transposed convolutions improves the speed of training and inference up to 36.3 and 5.1 times, respectively.

\section*{Acknowledgments}

Publication of this article has been funded by the Basic Science Research Program through the National Research Foundation of Korea (2018R1A5A1060031).

\bibliographystyle{siam}
\bibliography{paper}


\end{document}